\documentclass[conference]{IEEEtran}
\IEEEoverridecommandlockouts
\usepackage{cite}
\usepackage{amsmath,amssymb,amsfonts}
\usepackage{algorithmic}
\usepackage{graphicx}
\usepackage{textcomp}
\usepackage{xcolor}
\usepackage{svg}
\usepackage{tikz}
\usepackage{textcomp}
\usepackage{lipsum}
\usepackage[hidelinks]{hyperref}
\def\BibTeX{{\rm B\kern-.05em{\sc i\kern-.025em b}\kern-.08emT\kern-.1667em\lower.7ex\hbox{E}\kern-.125emX}}

\newcommand\copyrighttext{%
  \footnotesize \textcopyright 2023 IEEE. Personal use of this material is permitted.
  Permission from IEEE must be obtained for all other uses, in any current or future
  media, including reprinting/republishing this material for advertising or promotional
  purposes, creating new collective works, for resale or redistribution to servers or
  lists, or reuse of any copyrighted component of this work in other works.}
\newcommand\copyrightnotice{%
\begin{tikzpicture}[remember picture,overlay]
\node[anchor=south,yshift=10pt] at (current page.south) {\fbox{\parbox{\dimexpr\textwidth-\fboxsep-\fboxrule\relax}{\copyrighttext}}};
\end{tikzpicture}%
}
\begin{document}
\title{Safe Autonomous Driving in Adverse Weather: Sensor Evaluation and Performance Monitoring\\
\thanks{All authors are with the Institute of Safety in Future Mobility (ISAFE) of CARISSMA, Technische Hochschule Ingolstadt, Ingolstadt, Germany. Email: \texttt{\{fatih.sezgin, dagmar.steinhauser, daniel.vriesman, robert.lugner, thomas.brandmeier\}@carissma.eu.}}}

\author{\IEEEauthorblockN{Fatih Sezgin, Daniel Vriesman, Dagmar Steinhauser, Robert Lugner and Thomas Brandmeier}}


\maketitle
\copyrightnotice
\begin{abstract}
The vehicle's perception sensors radar, lidar and camera, which must work continuously and without restriction, especially with regard to automated/autonomous driving, can lose performance due to unfavourable weather conditions. This paper analyzes the sensor signals of these three sensor technologies under rain and fog as well as day and night. A data set of a driving test vehicle as an object target under different weather conditions was recorded in a controlled environment with adjustable, defined, and reproducible weather conditions. Based on the sensor performance evaluation, a method has been developed to detect sensor degradation, including determining the affected data areas and estimating how severe they are. Through this sensor monitoring, measures can be taken in subsequent algorithms to reduce the influences or to take them into account in safety and assistance systems to avoid malfunctions.
\end{abstract}


\section{Introduction}
Safety and assistance systems in vehicles rely heavily on the vehicle's perception sensors, such as radar, lidar, and camera, which must therefore operate continuously and without restriction to avoid failures in the triggering of restraint systems and the control of actuators. The need for reliable and robust sensor information will become even more important regarding autonomous driving. However, there are influences from adverse weather conditions that can reduce the performance of the sensors. The safety and assistance systems need to know if such an event occurs and if the sensors are affected by adverse weather conditions. Information about sensor disturbances is necessary to avoid malfunctions and reduce these influences. The effect of inaccurate or reduced sensor information for vehicle safety is discussed in \cite{CAVS_RL}, investigating sensor tolerances in the context of activation of irreversible restraint systems.

Several studies, such as \cite{WeatherConditionsonAutonomousVehicles, VriesmanRegen, Arage2006EffectsOW, rainpedestrian, WeatherInfluenceLidar, CameraSeeRain, RaindropsWindshield, weatherdetection, Goelles}, show that the different forward-looking sensors can be disturbed by various types of weather effects. \cite{WeatherConditionsonAutonomousVehicles} summarizes the impact of adverse weather conditions on state-of-the-art sensors and further investigates the effects of different weather conditions on autonomous driving systems using simulations. For heavy rain, an experimental study for radar and lidar based on artificially generated rain is presented in \cite{VriesmanRegen}, showing that water droplets affect the received information from objects in a static setting. Besides water droplets in the air, a layer of water on the radome can also degrade the signal of radar sensors \cite{Arage2006EffectsOW}. Steinhauser et al. discuss further rain-induced disturbances for pedestrian detection, emphasizing the importance of detecting bad weather \cite{rainpedestrian}. In \cite{WeatherInfluenceLidar}, the performance of a lidar sensor in rain and fog is analyzed, concluding that the number and the variance of the intensities of the received data points decrease. For the usage of camera systems under rain, \cite{CameraSeeRain} shows that rain provides a high variance in pixel intensity, which can be improved for certain cases by adjusting the camera parameters. Hasirlioglu et al. investigate the performance of object detection algorithms, such as YoloV3, in rainy conditions when raindrops are concentrated on the windscreen \cite{RaindropsWindshield}. As the accumulation of raindrops increases, the algorithm's detection performance decreases.
If the sensors do not function as intended, the fault must be detected and corrected to ensure safe operation. \cite{weatherdetection} presents a classification method for weather types based on camera and lidar. Goelles et al. present a review of FDIIR (Fault Detection, Isolation, Identification, and Recovery) methods for automotive perception sensors and a literature review for lidar sensors \cite{Goelles}. They mention that the most frequently reported errors in the literature are due to adverse weather conditions such as rain, fog, and snow, underlining the importance of researching these effects. They also conclude that detection and recovery methods for these faults still need to be included.
\\The motivation of the work presented in this paper is to continue existing research and show complementary aspects of weather effects on current sensors in the field of autonomous driving and integral vehicle safety. This results in deriving relevant parameters for performance evaluation and developing a monitoring system. Based on a self-diagnosis of the sensor measurements, this monitoring system continuously outputs whether, how strongly and in which regions the sensors are affected by weather influences. The instantaneous sensor performance evaluation sets this methodology apart from a sole weather classification and is an approach towards a recovery method.\\
The following section describes the sensor setup, the experimental design and the data processing. Section three presents the evaluation of the sensor data concerning weather effects. The fourth section covers the sensor monitoring approach, describes how it was developed and shows the results. The last section concludes the paper and addresses future work.

\section{Materials and Methods}
\label{sec:matmethods}
The required data were collected in the indoor test facility of the C-ISAFE Institute in the CARISSMA Research and Test Center of the Technische Hochschule Ingolstadt using its rain and fog system. The rain system covers an area of 50\,m x 4\,m and provides a controlled environment with reliable and reproducible rain conditions regarding intensity (from 16\,mm/h to 98\,mm/h), density, and droplet diameter. The generated fog is water-based, covers the test hall, and provides realistic conditions with adjustable visibility. 

\subsection{Measurement Setup}
\label{sec:Measurement Setup}
The measurements were made using three different sensor types: lidar, radar, and camera. The radar sensor, RadarLog from INRAS, was a chirp-sequence modulated FMCW radar with 16 receive antennas and raw data access. The centre frequency was set to 76.75\,GHz and the bandwidth to 1.5\,GHz resulting in a range resolution of 0.1\,m. The sampling frequency was 10\,MHz, the chirp duration 51.2\,$\mu$s and the repetition interval 60\,$\mu$s. An OS1 lidar from Ouster with 128 lines captured the scene with a frame rate of 20\,FPS. A 16-bit camera from PCO (model: Edge) with a resolution of 2560x2160 pixels and a set frame rate of 45\,FPS was used. These sensors were attached to a stationary sensor setup. The camera was mounted in a position corresponding to the position of camera systems behind the upper area of the windscreen. Radar and lidar were mounted at a height of 0.75\,m. The test vehicle was tracked with an indoor positioning system to obtain the ground truth position and velocity and easily distinguish vehicle detections from others during data analysis. The perception sensors and the indoor positioning system are synchronized using Unix-Timestamps and extrinsically calibrated to project the data onto each other during data processing. The test vehicle was a compact SUV (Audi Q2) with a white colour finish.

\subsection{Experimental Design}
The data was recorded with different scenarios and light/weather conditions. The weather conditions were dry (without rain or fog), fog (visibility of app. 8\,m) and rain with light (16\,mm/h) and heavy (98\,mm/h) rain intensities (in figures, this is abbreviated as L.Rain for the light and H.Rain for the heavy rain). The measurements were performed during day and night lighting condition. For the latter, the test facility was completely darkened. Measurements were performed in which the test vehicle drove centrally towards the sensor setup from a distance of 20\,m at velocities between 15 and 20\,kph. In addition, measurements were carried out without a test vehicle to capture the weather effects solely. The relevant region of interest covers an area of app. 20\,m x 4\,m in front of the sensors, correlating with the rain area.

\subsection{Data Processing}
\label{sec:Data Processingn}
\begin{figure}[tbh]
    \centering
    \includegraphics[width=0.92\columnwidth]{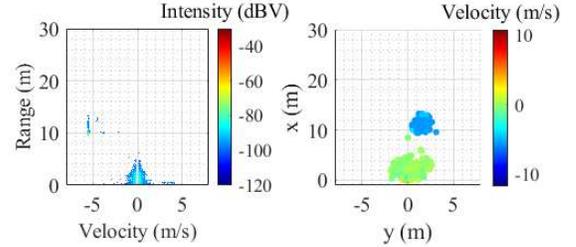}
    \caption{Radar detections under light rain with a vehicle located at a distance of approx. 10\,m. The detections up to 5\,m are detections generated by rain. Left: Range-Doppler map. Right: Detections in Cartesian coordinates.}
    \label{fig:RDandXYZ}
\end{figure}

After the electromagnetic wave emitted by the radar reaches an object, it is reflected, received, mixed, filtered and sampled. The first step in processing the data, i.e. generating the detection points, is to perform a two-dimensional FFT (Fast Fourier Transformation) on the sampled time signal to provide the range-Doppler information. Then OS-CFAR (Ordered Statistic Constant False Alarm Rate) filtering is applied to the range-Doppler spectra to remove the noise. Fig.\,\ref{fig:RDandXYZ} shows the resulting range-Doppler map on the left. The Burg method is used to obtain the angular information \cite{Burg, SezginPed, Held}. The result is depicted in Fig.\,\ref{fig:RDandXYZ} on the right.

The raw data were used as point clouds for the lidar and PNG (Portable Network Graphic) images for the camera. The position of the test vehicle was labeled using the data received from the indoor positioning system. The labeling was first done with the lidar data. In this step, the position information is transformed into the lidar coordinate system and 3D bounding boxes are automatically generated for each frame. The bounding boxes are then projected onto the radar and camera frames using the extrinsic calibration parameters for both sensors.

\section{Weather Effects on Sensor Data}
\label{sec:WEATHER EFFECT ON SENSOR DATA}
This section discusses the effects on the data of the different sensor types that are attributable to the different weather conditions.

\subsection{Radar}
\begin{figure}[tbh]
    \centering
    \includegraphics[width=0.92\columnwidth]{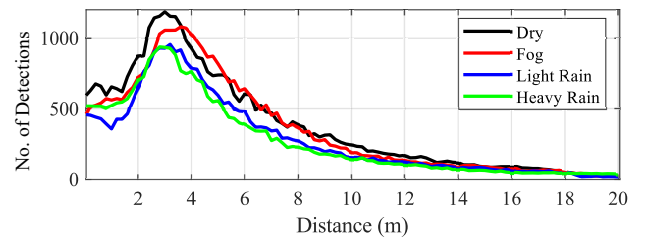}
    \caption{Number of detections originating by the test vehicle.}
    \label{fig:NrDetPoints}
\end{figure}

First, the weather effects are considered with regard to the detections reflected by the target vehicle. The number of vehicle detections dependent on distance in the four weather conditions is compared in scenarios representing a test vehicle approaching the sensor stand. Fig.\,\ref{fig:NrDetPoints} shows the respective curves. Each of these are obtained through averaging four measurements that were recorded with similar driving paths and velocities.

\begin{figure}[tbh]
    \centering
    \includegraphics[width=0.92\columnwidth]{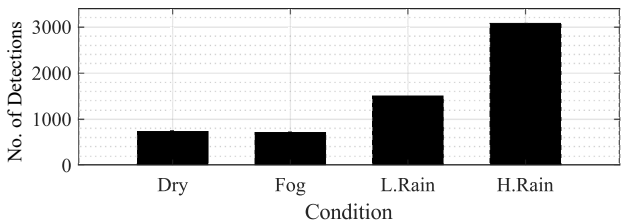}      
    \caption{Size of the nearest cluster to the sensor in the range-Doppler map.}
    \label{fig:SizeCluster}
\end{figure}

It can be observed that despite adverse weather conditions, the radar can detect the object with a high number of data points under all tested conditions. Nonetheless, most detection points were captured in dry conditions and the least in heavy rain. A slightly lower number can also be seen in light rain and fog. One reason for this could be the formation of droplets or a layer of water on the radome. Since the recorded position is the center of the vehicle, the number of detection points decreases after a certain distance as the vehicle passes the sensor stand.
\\The size of the cluster located directly in front of the sensor is extracted from the range-Doppler data. The cluster in question extends centrally around the velocity 0\,m/s over the entire range. All static detections are always located in this cluster. In the example plot in fig.\,\ref{fig:RDandXYZ} on the left, additional detections are located close to the sensors. These originate from rain droplets reflecting the signal to the sensor. The clustering algorithm adds these points to the cluster of the static objects and therefore extends the velocity in this cluster up to approx. 5\,m. This changes the number of occupied cells and the cluster size, as seen in fig.\,\ref{fig:SizeCluster}. The cluster in light rain is smaller than that in heavy rain but larger than in the frames taken in fog and dry conditions. The clusters in the frames taken during fog and dry are the smallest and at a similar level.

\begin{figure}[tbh]
    \centering
    \includegraphics[width=0.92\columnwidth]{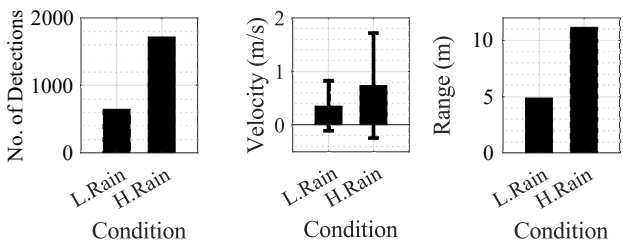}      
    \caption{Comparisons of various rain effects captured by the radar.}
    \label{fig:RadarEffects}
\end{figure}

The number of detections generated by water droplets in heavy rain is significantly higher than in light rain, cf. Fig.\,\ref{fig:RadarEffects} on the left. The middle chart compares the mean velocity and standard deviation. Light rain has lower velocities, and the standard distribution is also lower than in heavy rain. On the right, it can be seen how far the radar detects the rain at these intensities. In light rain, this value is about 5\,m and in heavy rain about 11\,m. The charts in Fig.\,\ref{fig:RadarEffects} show the average of 50 radar frames in each bar. 
\\In summary, the radar is generally robust in challenging weather conditions. However, water droplets in the air are detected and can cause interference in the radar frames.

\subsection{Lidar}
\begin{figure}[tbh]
    \centering
    \includegraphics[width=0.9\columnwidth]{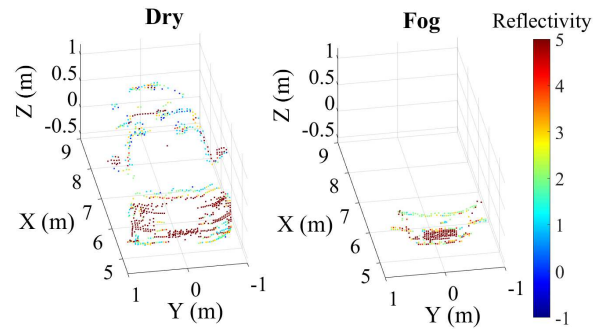}      
    \caption{Appearance of the test vehicle in lidar data at 5\,m distance in dry and foggy condition. Target reflectivity is obtained through Sensor Signal Photons measurements, scaled on measured range and sensor sensitivity at that range \cite{ousterUM}.}
    \label{fig:AppLidar}
\end{figure}

Fig.\,\ref{fig:AppLidar} shows examples of how the test vehicle appears in dry and in fog. In comparison, it can be clearly seen that there is a significant loss of information in the latter. Vriesman et al. use dispersion as a metric to analyze information loss in rain for vehicle detections in lidar frames with static scenarios \cite{VriesmanRegen}. This metric is also used here for the comparisons of weather effects. The dispersion of the detection points within the vehicle bounding box corresponds to the variance of the distances of these points $P$ to the origin point $O$ of the bounding box in eq.\,\ref{eq:disp}. 

\begin{equation}
    \mathrm{dispersion} = \sigma^{2}(d(O, P))
    \label{eq:disp}
\end{equation}

\begin{figure}[tbh]
    \centering
    \includegraphics[width=0.92\columnwidth]{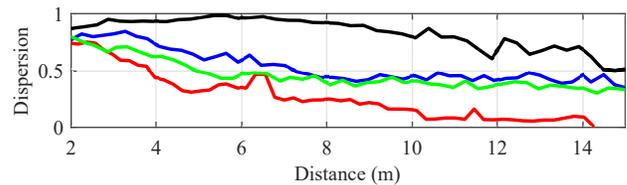} 
    \caption{Comparison of the dispersion of the vehicle detections in different weather conditions. Black: Dry, red: Fog, blue: L.Rain, green: H.Rain.}    \label{fig:DispLidar}
\end{figure}

Fig.\,\ref{fig:DispLidar} compares the dispersion values within the vehicle bounding boxes in measurement drives under different weather conditions. Each curve was also determined by means of four measurement drives. When it rains, a significant loss of information is already recorded. Compared to radar, lidar is significantly more affected by rain. The loss of information becomes even more apparent in fog.

\begin{figure}[tbh]
    \centering
    \includegraphics[width=0.92\columnwidth]{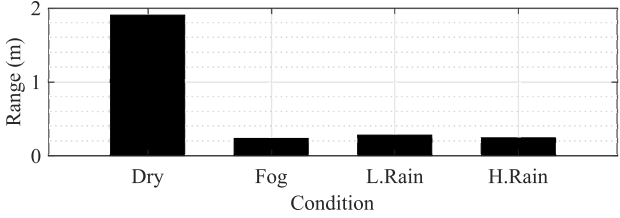}     
    \caption{Distance to the nearest lidar detection in different weather conditions.}
    \label{fig:DistLidar}
\end{figure}

Up to this point, the information losses regarding the test vehicle have been presented. Next, the entire relevant scene captured by the lidar is considered. The smallest distance at which the lidar outputs a detection point is shown in Fig.\,\ref{fig:DistLidar}. In dry conditions, the ground is typically detected first, with the distance to the first detection depending on the installation height of the sensor. In adverse weather conditions the minimum distances of the detection points are very low, as the lidar detects droplets in the air. A similar observation is also made in dense fog. The lidar detects the fog aerosols at a very short distance. When fog envelops the sensor, it causes the ring phenomenon, resulting in a doughnut-shaped pattern of detection points around the lidar \cite{lidarring}. 

%

\begin{figure}[tbh]
    \centering
    \includegraphics[width=0.92\columnwidth]{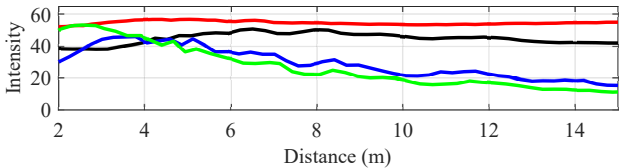}      
    \caption{Mean intensity of the lidar detections with an driving vehicle in the scene. Black: Dry, red: Fog, blue: L.Rain, green: H.Rain.}
    \label{fig:IntLidar}
\end{figure}

Fig.\,\ref{fig:IntLidar} shows the different mean intensities in the lidar frames over the vehicle distance of an approaching vehicle. These curves consider the intensities of all detection points in the relevant field of view. The ring phenomenon described earlier explains that the highest values are constantly in foggy conditions. The curves of the rainy conditions have a similar level throughout. For both, the value increases as the vehicle's distance decreases. Raindrops in the air cause signal attenuation, resulting in lower mean intensity of detections at higher distances. 

Impacts of weather effects on ground detections were also investigated. Ground detections are lower in adverse weather conditions than in dry conditions. In the latter's case, there are ca. 14,000 detections in the relevant field of view. Fog and rain reduce detections to ca. 2,000 and ca. 8,000, respectively. These are values averaged over 50 frames. One reason for the reduction of ground detections in fog is the aforementioned ring phenomenon, which prevents the rays from reaching the more distant ground. In rain, puddles form on the ground, deflecting the rays. 
\\In summary, the lidar performance suffers greatly under adverse weather conditions. Even light rain leads to information losses and lower intensities. With increasing rain intensities, these effects become more pronounced. The strongest degradations are observed in dense fog.


\subsection{Camera}
\begin{figure}[tbh]
    \centering
    \includegraphics[width=0.92\columnwidth]{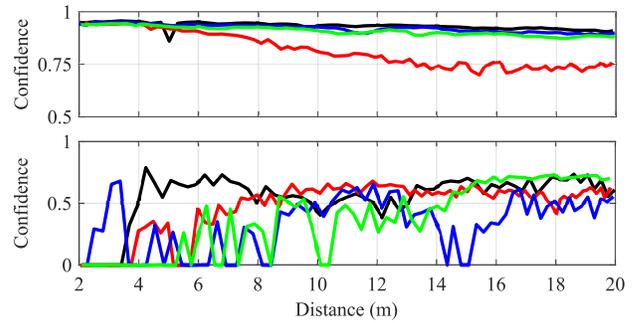}
    \caption{Confidence scores of vehicle detection of YoloV5 at daytime (upper) and nighttime (lower). Black: Dry, red: Fog, blue: L.Rain, green: H.Rain.}
    \label{fig:YoloV5ConfDay}
\end{figure}

As part of the analysis of weather effects on the camera, in addition to other metrics such as sharpness, brightness, and contrast in the frames, the YoloV5 \cite{YoloV5} Object Detector was considered to see how the performance of such algorithms is affected by the weather effects captured in the camera images. 
For this purpose, the model was trained with the BDD100K dataset \cite{BDD100K}, which contains relevant data related to the scenario, light, and weather conditions, from measurements in real road traffic. 
It was then applied to the collected data described above. The detection quality in the form of the confidence score of the vehicle depending on its position is investigated. In Fig.\,\ref{fig:YoloV5ConfDay} upper plot, the comparison of these can be made under different weather conditions during daylight. It is noticeable how much the fog affects the performance of the detector. Only at short distances, the object values reach a level similar to that in dry and rainy conditions. Fig.\,\ref{fig:YoloV5ConfDay} lower plot shows the object scores in different weather conditions at night. These are significantly below the performance during the day, and the vehicle is often not detected, although nearly half of the training data represent night scenarios. However, no light was shining in the direction of the target vehicle during the tests, and only the vehicle lights were on.

\begin{figure}[ht!]
    \centering
    \includegraphics[width=0.92\columnwidth]{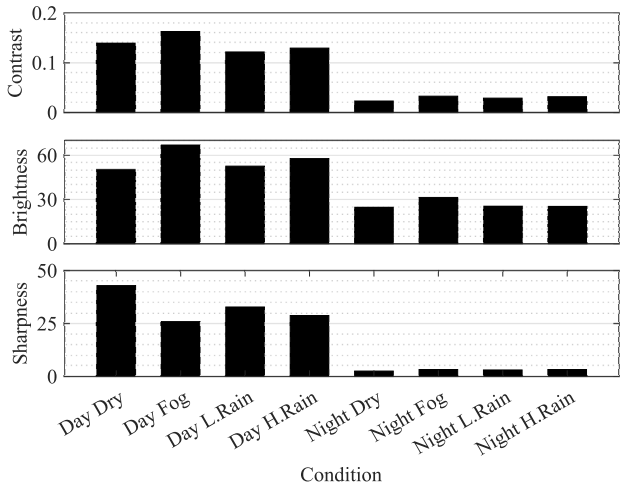}      
    \caption{Sharpness, brightness, and contrast in the images in different conditions calculated with Laplace operation.}
    \label{fig:Sharpness}
\end{figure}

The metrics of sharpness, brightness, and contrast were also considered in the analysis of camera degradation (Fig.\,\ref{fig:Sharpness}). For each evaluation, a measurement with approximately 150 camera frames was used. The sharpness was calculated using a Laplace operation \cite{blur_detection}.
The blurriest shots were those taken in night conditions. One reason for this could be the longer exposure time automatically set by the camera.
The most eminent blurriness in daylight was produced in fog, the least in dry conditions. In total, all the weather conditions have a negative influence on this value. Besides this, it is shown that the light conditions play a vital role in the detection performance. This should always be taken into account. The brightness of the frames was determined by transforming them into the HSV color space and observing the value. 
The brightest frames were recorded in daylight fog. For the night scenarios, the brightest frames were also captured in fog. One reason is that the fog particles on a large part of the frame scatter the vehicle's light. As expected, the frames recorded during night-time scenarios were generally darker than during daytime.

The Root-Mean-Square (RMS) contrast \cite{rms_contrast} is used to compare the image contrasts for the different weather conditions in Fig.\,\ref{fig:Sharpness}. The contrast in the frames under daytime conditions is higher than that of the frames in the dark. The contrast is highest for the bright and foggy frames. The image histograms illustrate that the fog distributes the image intensities. Thus, the intensity range in the fog is greater than that in the dry at this specific recording site.
\\Consequently, the camera is affected by poor lighting conditions and the adverse weather conditions. Even light rain causes changes in the investigated parameters. As with the lidar, fog causes the strongest influences.


\section{Sensor Monitoring}
\label{sec:SENSOR MONITORING}
After analyzing the data regarding weather influences, a monitoring system was set up for the sensors to determine their data quality in different weather conditions automatically. This section describes the underlying approach. A sensor monitoring system based on a fuzzy logic tree and genetic algorithm \cite{geneticFuzzy} was developed. One reason for choosing fuzzy logic over black box methods, such as neural networks, is the requirement in vehicle safety to use controllable and transparent methods. First, a development data set is built. Parameters and information affected by the weather are extracted from the data frames. The chosen parameters for the radar include the standard deviation of velocity, the mean and standard deviation of the point cloud density calculated for all data points with a spherical radius of one meter, the intensities in the raw data, and the cluster sizes determined in the whole frame and separately only for the point closest to the sensor. The values for point-cloud density and intensities and the changes in maximum and minimum distances are extracted for the lidar. For the camera, contrast values, sharpness, and brightness are used. Since the rules are built and optimized using genetic algorithms, methods for consistent labeling of the performance of the sensors under different weather conditions are needed. The following is used for the different sensors: the dispersion in the vehicle bounding for lidar, the object confidence score of YoloV5, and additionally, the intersection of unions (IOU) with the ground truth bounding box for the camera and the number of detection points of the test vehicle for radar. These values were translated with thresholds into the expressions good, poor, and moderate and added to the extracted information. The data set thus includes the data frame, the extracted information that changes with different weather conditions, and the labeled sensor performance determined. With this data set, the fuzzy inference systems (FIS) in the fuzzy trees were constructed and optimized using a genetic algorithm. Each FIS has two inputs and one output. The sensor performance is output at the end of the entire fuzzy tree. For radar and lidar, a fuzzy tree is set up for each 2-meter distance range since the sensor performance depends on the distance.

\begin{figure}[tbh]
    \centering
    \includegraphics[width=0.75\columnwidth]{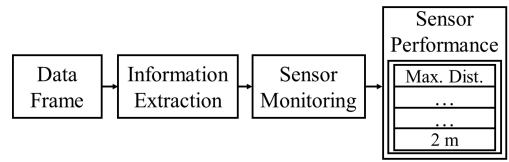}   
    \vspace{-0.5em}     
    \caption{Monitoring workflow.}
    \label{fig:MonitoringRadar}
\end{figure}

The sensor monitoring module is developed and tested with new data. Fig.\,\ref{fig:MonitoringRadar} presents a workflow for automatically evaluating sensor performance when applied to frames. The process starts by feeding a data frame into the system. The system then extracts the information from the frame and transfers it to the sensor monitoring module. The module outputs the sensor performance based on the fuzzy trees that were previously built. Additionally, the module determines the grid maps for the disturbed velocity and spatial areas for the radar, using the nearest cluster calculated from the Doppler data and the spatial information of the detection points. As demonstrated, this particular cluster includes rain information. The 3D clustering with the velocity information allows for separation even when the vehicle is inside the rain area. Moreover, the shape of the rain cluster in the range-Doppler data can indicate whether a target object is moving into it. For instance, the shape changes if the detected vehicle slows down at a short distance and enters the velocity range of the rain.

\begin{figure}[tbh]
    \centering
    \includegraphics[width=0.72\columnwidth]{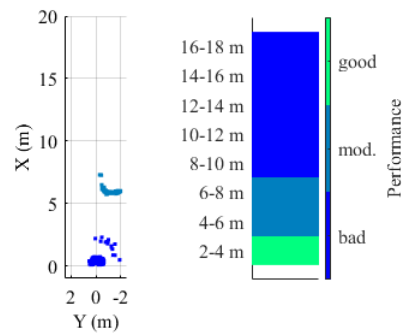}      
    \vspace{-1em}
    \caption{Monitoring of lidar in fog.}
    \label{fig:ErgLidar}
\end{figure}

The monitoring system underwent testing under distinct weather conditions and examples are given in the following. Fig.\,\ref{fig:ErgLidar} displays the result for the lidar frame on the left, which was captured in fog. The quality of the frame at a close range received a good rating, but it decreases as the distance increases.

\begin{figure}[tbh]
    \centering
    \includegraphics[width=0.5\columnwidth]{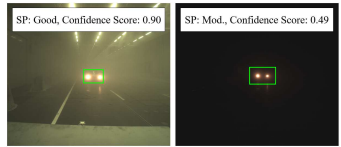}      
    \caption{Monitoring of camera in fog.}
    \label{fig:ErgCamera}
\end{figure}

Fig.\,\ref{fig:ErgCamera} displays camera frames taken in fog under varying lighting conditions. The quality is good for the daytime frame and moderate for the nighttime frame. For comparison, the detection bounding box and the confidence value of YoloV5 for the detected object are also shown. On the left frame, the bounding box closely matches the size of the detected vehicle, while on the right frame, the bounding box is larger than the object.

\begin{figure}[tbh]
    \centering
    \includegraphics[width=0.45\columnwidth]{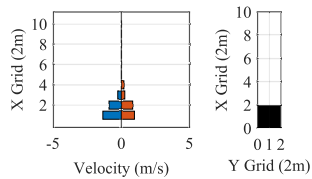}   
    \caption{Monitoring result of radar in light rain.}
    \label{fig:ErgRadar}
\end{figure}

In all test drives, the radar is a robust sensor in various weather conditions, detecting the vehicle with ample information even in heavy rain. However, the radar also detects raindrops, which appear at close ranges. To account for this interference, we created grid maps as described above, indicating which velocity and spatial areas are impacted. Fig.\,\ref{fig:ErgRadar} displays the result of the example under light rain from Fig.\,\ref{fig:RDandXYZ}.

\section{Conclusions}
\label{sec:conclusions}
In summary, the analysis of sensor performance presented in this paper has shown that radar, camera, and lidar can be affected by adverse weather, such as rain and fog. In addition, the lighting conditions play a major role in the performance of cameras. This highlights the importance of considering such phenomena when developing safety, assistance, and autonomous driving functions. Furthermore, an approach for a sensor monitoring system was presented. The proposed solution based on fuzzy logic and genetic algorithms to monitor sensors have the potential to evaluate and output the reliability of sensor systems, which is an important step towards maintaining optimal performance for sensor systems applicable for safe automated and autonomous driving. More research is needed to fully explore the possibilities of this approach and test its performance in real-world scenarios.

\section*{Acknowledgments}
The authors would like to thank the test engineers Christopher Ruzok and Michael Graf for their support in conducting the measurements, as well as e:fs TechHub GmbH for their support. This work was conducted within the projects SAVE-ROAD (Bavarian Research Foundation, grant number AZ-1355-18) and KICSAFe (Bayerisches Staatsministerium für Wirtschaft, Landesentwicklung und Energie, grant number DIK0461/01).

\bibliography{bibfile.bib}{}
\bibliographystyle{ieeetr}

\end{document}